\documentclass[sigconf]{acmart}
\usepackage{graphicx}
\usepackage{multirow}
\usepackage{capt-of}
\usepackage{pdflscape}
\usepackage{afterpage}
\usepackage{color}
\usepackage{balance} 
\usepackage{booktabs}
\usepackage{url}

\acmConference[KDD-MLMH2018]{KDD Workshop on Machine Learning for Medicine and Healthcare}{Aug 2018}{London, United Kingdom} 
\acmYear{2018}
\copyrightyear{2018}

\begin{document}
%
\title{A hybrid deep learning approach for medical relation extraction}

\author{Veera Raghavendra Chikka}
\affiliation{%
 \institution{International Institute of Information Technology}
 \streetaddress{Gachibowli}
 \city{Hyderabad} 
 \country{India}}
\email{raghavendra.ch@research.iiit.ac.in}

\author{Kamalakar Karlapalem} 
\affiliation{%
 \institution{International Institute of Information Technology}
 \streetaddress{Gachibowli}
 \city{Hyderabad}
\country{India}}
\email{kamal@iiit.ac.in}

\begin{abstract}
Mining relationships between treatment(s) and medical problem(s) is vital in the biomedical domain. This helps in various applications, such as decision support system, safety surveillance, and new treatment discovery. We propose a deep learning approach that utilizes both word level and sentence-level representations to extract the relationships between treatment and problem. While deep learning techniques demand a large amount of data for training, we make use of a rule-based system particularly for relationship classes with fewer samples. Our final relations are derived by jointly combining the results from deep learning and rule-based models. Our system achieved a promising performance on the relationship classes of I2b2 2010 relation extraction task.

\end{abstract}

\keywords{Medical Relation Extraction, Bi-directional LSTM, Deep Learning}

\maketitle
%

\section{Introduction}

Understanding the relationships between treatment and problem is crucial in various biomedical applications. The extensive generation of electronic health records prompt immense knowledge of these relations in the biomedical literature. Manually extracting these relations is difficult and time-consuming. This necessitates the automatic techniques with minimal human intervention. Various challenges have been organized in the past decade to identify relationships in the biomedical literature~\cite{uzuner20112010}. Relation extraction usually includes two steps: (1) identifying named entities, (2) identifying relations between them. Unlike named entities, relation extraction is a complex task which aims to extract context and structure from the text \cite{ananiadou2010event}. A simple bag-of-words and co-occurrence modelling cannot detect the relationship among these entities. Studies~\cite{chun2006extraction} report that only 30\% of co-occurring protein-protein entities have actual interactions. Also, relationships have internal structures which are very difficult to detect~\cite{ananiadou2010event}.

In order to promote the research in relation extraction, I2b2 2010 community organized a challenge~\cite{uzuner20112010} on identifying relationship types between medical problems, and treatments. 
The relationships between treatment and problem addressed in this task are:
\begin{enumerate}
\item Treatment is administered for a medical problem (TrAP), suggests that this treatment is an accepted procedure for the medical problem. For example, ``She did have some pain and was treated with Percocet''.
\item Treatment improves the medical problem(TrIP). For example, ``He had occasional episodes of rapid atrial fibrillation that were controlled with IV metoprolol''.
\item Treatment worsens the medical problem (TrWP). For example, ``The stroke occurred while taking aspirin and plavix''.
\item Treatment causes medical problem(TrCP). For example, ``The patient had pancytopenia and vomiting on DDI''.
\item Treatment is not administered because of a medical problem (TrNAP), suggests that this treatment should not be used for this medical problem. For example, ``His coumadin was held during his stay given his acute bleed''.
\end{enumerate}

Most of the participants in I2b2 2010 challenge have used machine learning techniques like Support vector machines (SVM) for the identification of relationship type among the entities~\cite{de2010nrc}\cite{jiang2010hybrid}\cite{patrick2010i2b2}\cite{roberts2010extraction}. Among the participant systems, hybrid approach \cite{roberts2010extraction} with SVM and rule-based approach reported top performance in the relation extraction task. 
Our goal is to develop a hybrid method to improve the performance of deep learning technique. The contributions of our work are: 
\begin{itemize}
\item Usually CNN is the popular neural network architecture used in relation extraction task. In this work, we have experimented with Bi-directional Long Short Term Memory (LSTM) to capture sequential dependency between the relation arguments.
\item Comparison and analysis of SVM and deep learning for relation extraction task have been reported.
\item For the task of relation classification between the medical entities, the state of art approaches consider only the word-level and three word window proximity. We experiment by considering the complete sentence as a matrix comprising word vectors of all the words along with position indexes.
\end{itemize}

\section{Bi-directional LSTM}

Long-Short Term Memory (LSTM) is an improvement over Recurrent Neural Networks (RNNs) designed to overcome the problem of exponential decay of back propagation error~\cite{hochreiter1997long}. Thus enabling LSTMs to handle long-term dependencies. LSTM layer is a chain-like structure of components, known as memory blocks. Each block contains one or more memory cells and three regulatory structures - input, output and forget gates.
These gates allow LSTM to add, delete, and reset the information flowing through each block and stores the information in the form of cell state~\cite{yu2015using}. 

\begin{figure}
\centering
	\includegraphics[height=9.5cm, width=0.45\textwidth]{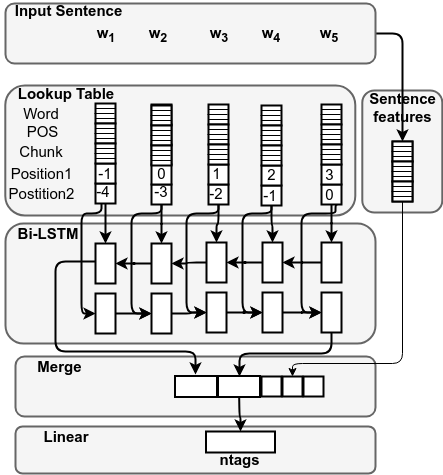}
\caption{Bi-directional LSTM}
\label{fig:brnn}
\end{figure}

LSTMs flows the data in one directional, where as Bidirectional-LSTM processes the information from both directions. Figure \ref{fig:brnn} shows the LSTM hidden layer comprising of LSTM block passing information from either side of the layer. This allows LSTM to learn the contextual dependency of current word on either sides. 
Figure~\ref{fig:brnn} shows our neural network architecture for relation extraction task. Input sentence is passed for each sample pair of treatment and problem present in the sentence. The sentence is represented with both word-level and sentence level features. Word level features
used are: word token, Part of speech tags, and Chunk phrase tags. Lookup table computes the representations for each of the features along with position vectors. Position vector contains the distance of the word from the treatment and problem for which relationship type has to be identified~\cite{collobert2011natural}.  
For example, if $w_2$ and $w_5$ are the medical entities, then the position vectors of the sentence for entities $w_2$ and $w_5$ are (-1, 0, 1, 2, 3) and (-4, -3, -2, -1, 0), respectively. The features are initialized with the random representations. These representation are fine-tuned while training through back propagation.
\subsection{Sentence level features for relation extraction}
\label{additionalfeatures}
\begin{enumerate}
\item POS tag sequence: The part-of-speech sequence of the phrase between the entities. Frequent 100 POS tag sequence are pre-computed from the training dataset. The vector of dimension 100 is used as feature.
\item Point wise mutual information: A statistical measure between two entities, based on their co-occurrences. This value is used as a feature.
\item Added feature words used to find the assertion of the sentence. These are the dictionary keywords used in Apache cTAKES~\cite{savova2010mayo} (namely, allergy.txt, cause.txt, fail.txt, certainty.txt, history.txt, hypothetical\_cue\_list.txt and uncertainty.txt). These words give the context of the sentence. The index of the word tokens in each of the dictionary is used as feature.
\end{enumerate}

Sentence level representations are computed by a separate module. These representations are appended Bi-LSTM outputs in the Merge Layer. Finally, a fully connected Linear layer is used to get the output size of number of relationships (ntags).


\section{Rule based approach}
Rule-based approaches are widely used in medical applications which need critical quality for high precision outcomes. Relationship between pair of medical terms depends on the context in which they occur in the sentence.
For example, a sentence $s = w_1, ..., m_1, ..., w_i , ..., m_2, ..., w_n$ containing medical terms $m_1$, and $m_2$ with relation $r$ can be described as $s = s_b m_1 s_m m_2 s_a$, where $s_b$, $s_m$ and $s_a$ are before, middle and after word-context portions of the medical terms ($m_1$, and $m_2$) respectively. 

Given a test sentence $t$ containing medical terms $m_{01}$ and $m_{02}$, if the before, middle and after portions of $t$ are similar to that of sentence $s$, it implies that the relation $r$ of $s$ exists in between $m_{01}$ and $m_{02}$. Two approaches were used to identify the relation ship between medical terms:
\vspace{-2.0em}
\begin{itemize}
\item \textbf{Pattern of sentence/phrase:} The relation of two entities in a sentence depends on the phrase in between the entities. Such phrase patterns were manually framed by checking the training dataset samples. For example, a pattern for relation ``TrAP" (Treatment Administered Problem) can be defined as \\``$<problem>$ is diagnosed with $<treatment>$".\\
\item \textbf{Shortest Dependency path:} The verbs that are present in the shortest dependency path of pair of the medical terms can be used to identify the relation. For example, \\
\textbf{Sentence:} Given her \textit{fever} the patient was treated with \textit{Ceptaz} and \textit{Levaquin}.\\
\textbf{Medical entities:} fever (problem), Ceptaz (drug), and Levaquin (drug) \\
\textbf{Relationships:} (Ceptaz, fever, TrAP), (Levaquin, fever, TrAP). \\
In the above sentence, the verb ``treated'' infers that the medicines ``Ceptaz'' and ``Levaquin'' were administered for problem ``fever''. Having verb ``treated" in the dependency path of pair of entities `Ceptaz' and `fever', infers the relationship TrAP. Similarly, for the pair `Levaquin' and `fever'. If there is no dependency path exists in between the medical terms, the string of words between the medical terms is used as the path. The verbs for each of the relation are collected from the training dataset. For the relationship 'TrCP' (Treatment controls problem), has verbs like control, regulate, modulate, restrict, etc., are used to identify the relationship. We check whether the shortest dependency path between the entities contains such terms.

\end{itemize}

\section{Results and Discussion}

\subsection{Dataset statistics:}
\label{sec:datastats}
We used a subset of i2b22010 challenge dataset which is publicly available. The organizers of I2b2 have restricted the distribution of remaining dataset. The original training dataset consists of 349 documents, 27,837 concepts, and 5264 relations, and the test set consists of 477 documents, 45,009 concepts, and 9069 relations. Statistics of our i2b2 2010 subset data is given in Table~\ref{tab:i2b22010stats_rel}. Out of the available 426 documents, 10\% (42 documents) were used for testing, and the remaining 384 documents were used for training.
This was done in order to maximize the training set to provide as large as possible dataset for deep learning purposes. All the experiments are carried out using this dataset. 

\begin{table}[htb]
\small
\begin{center}
\caption{Relationship type instances}
\begin{tabular}{|c|c|c|c|}
\hline
\multirow{2}{4em}{Relationship types}&\multicolumn{3}{|c|}{Number of instances}\\
&Total& Train & Test\\
\hline
TrAP &2617&2342&275\\
TrCP&526&455&71\\
TrIP&203&167&36\\
TrNAP &174&161&13\\
TrWP&133&119&14\\
All&3653&3244&409\\
\hline
\end{tabular}
\label{tab:i2b22010stats_rel}
\end{center}
\end{table}

\vspace{-0.5em}
\subsection{Relation Extraction Experimentation}
Experiments are evaluated using commonly used Precision (P), Recall (R) and F-score (F). These metrics are computed based on the True Positives (TP), False Positives (FP), and False Negatives (FN). \\
$Precision (P) = \frac{TP}{TP+FP}$\\\\
\vspace{0.1em}
$Recall (R)= \frac{TP}{TP+FN}$\\\\
\vspace{0.1em}
$F_{score}(F) = \frac{2 \times   Recall  \times Precision}{Recall + Precision}$\\

For the relation extraction task, we have used keras \cite{chollet2015keras} for the implementation of Bi-LSTM neural network architecture.  
As done in previous works, we train the relation extraction model by giving the gold relations and test the model by checking for relations between gold medical entities. The parameters of LSTM architecture that are restored to default are $LSTM\_output\_size$ as 64 and $number~of~epochs$ as 20. 

\subsubsection{Effect of negative samples:}
The total number of samples in our dataset with respect to relations is around 4000 as given in Table \ref{tab:i2b22010stats_rel}. In order to balance these positive samples, we have experimented with varied number of negative samples ($null$ relationships) as 5000, 10000, 20000 and 30000, where we have obtained higher F-score for 20000 negative samples. 

\subsubsection{Effect of Varying Embedding size:}
Word embedding is the numerical representation of words in the form of vectors. We used Glove tool \cite{pennington2014glove} to obtain vector representations (embeddings) for words. We have empirically tried various word embedding sizes. Results on embedding sizes 40, 100 and 200 are shown in Table \ref{tab:results_rel_emb}. Based on Table~\ref{tab:results_rel_emb}, we chose to use embedding size $40$ for further experimentation.

%

\begin{table*}
\caption{Relation Extraction Results with varying embedding sizes}
\begin{tabular}{|c||c|c|c|c|c|c||c|c|c|c|c|c||c|c|c|c|c|c|}
\hline
&\multicolumn{18}{|c|}{Embedding sizes} \\
\hline
&\multicolumn{6}{|c||}{40}& \multicolumn{6}{|c||}{100}& \multicolumn{6}{|c|}{200} \\
\hline
&\small{TrAP}&\small{TrCP}&\small{TrIP}&\small{TrNAP}&\small{TrWP}&\small{Total}
&\small{TrAP}&\small{TrCP}&\small{TrIP}&\small{TrNAP}&\small{TrWP}&\small{Total}
&\small{TrAP}&\small{TrCP}&\small{TrIP}&\small{TrNAP}&\small{TrWP}&\small{Total}\\
\hline
P& 0.53&0.44&0.5 & 0.45 &0.08&0.51 & 
0.563&0.48&0.26&0.17&0.08&0.45 & 
0.49& 0.42& 0.37&0.23&0.07& 0.44\\
R& 0.60 &0.30&0.25 &0.38 &0.07 &0.49 
&0.6 & 0.35&0.38&0.38&0.21&0.51
&0.61& 0.42&0.27&0.46&0.14& 0.51\\
F& 0.56&0.36 & \textbf{0.33} &\textbf{0.41}& 0.0&\textbf{0.50} 
&\textbf{0.58}&\textbf{0.40}&0.31&0.24&\textbf{0.12}&0.48 
&0.54&0.35&0.31&0.30&0.09&0.47\\
\hline
\end{tabular}
\label{tab:results_rel_emb}
\end{table*}

\begin{table*}[htb]
\caption{Relation Extraction Results using SVM (baseline), Bi-LSTM and Rule-Based approaches}
\begin{tabular}{|c||c|c|c|c|c|c||c|c|c|c|c|c||c|c|c|c|c|c|}

\hline
&\multicolumn{6}{|c||}{$SVM$ (baseline)}& \multicolumn{6}{|c||}{$Bi-LSTM$}& \multicolumn{6}{|c|}{$Rule Based$}\\
\hline
&\small{TrAP}&\small{TrCP}&\small{TrIP}&\small{TrNAP}&\small{TrWP}&\small{Total}
&\small{TrAP}&\small{TrCP}&\small{TrIP}&\small{TrNAP}&\small{TrWP}&\small{Total}
&\small{TrAP}&\small{TrCP}&\small{TrIP}&\small{TrNAP}&\small{TrWP}&\small{Total}\\
\hline
P& 0.36&0.65&0.61&0.26&0.33&0.38
&0.61& 0.47&0.41&0.14&0.15& 0.50 
&0.56 &1.0 &1.0 &1.0 &1.0 & 0.61 \\
R& 0.71&0.36&0.36&0.30&0.07&0.58
& 0.56&0.47&0.36&0.30&0.21&0.52
&0.29 & 0.09 &0.22 &0.23 &0.07 &0.24\\
F& 0.48&0.46&\textbf{0.45}&0.28 &0.11&0.46
& \textbf{0.58}&\textbf{0.47}&0.38&0.20&\textbf{0.17}&\textbf{0.51}
&0.38 &0.17 &0.36&\textbf{0.37}&0.13& 0.35\\
\hline
\end{tabular}

\label{tab:results_rel_additionalfeatures}
\end{table*}

\begin{table*}
\caption{Results of Hybrid Bi-LSTM and Rule-based system}
\begin{tabular}{|c|c|c|c|c|c|c|}
\hline
& \multicolumn{6}{|c|}{Bi-LSTM + RuleBased}\\
\hline
&TrAP&TrCP&TrIP&TrNAP&TrWP&Total \\
\hline
Precision 
&0.61& 0.47&0.44&0.22&0.15& 0.51 \\
Recall 
& 0.56&0.47&0.41&0.46&0.21&0.53 \\
F-score 
& \textbf{0.58}&\textbf{0.47}&\textbf{0.42}&\textbf{0.30}&\textbf{0.17}&\textbf{0.52}\\
\hline
\end{tabular}
\label{tab:sota-re}
\end{table*}



\subsubsection{Comparison of SVM, Bi-LSTM and Rule-based approaches}
Table \ref{tab:results_rel_additionalfeatures} shows the results of relation extraction with SVM, Bi-directional LSTM and Rule based approaches. 
From Table \ref{tab:results_rel_additionalfeatures}, we can infer that 
Bi-LSTM has shown relatively better performance on identifying TrAP relation, implying the importance of the large number of data samples required to train a neural network model. 

As the part of the I2b2 dataset is retricted to use, we have built a SVM model on remaining dataset. We have used libsvm \cite{chang2011libsvm} for the implementation of SVM based on the top performing system ~\cite{rink2011automatic} of the i2b2 challenge. The feature vector for SVM is given as a one-hot encoded vector of word vocabulary, part of speech tags, chunk tags and position indexes. SVM has shown a consistent results on identifying TrAP, TrCP, and TrIP which are around 0.45 F-score. This implies that the SVM model can be trained with comparatively less number of samples and it also balances the class weights of the relationship types.

Framing rules for relation extraction is a very challenging task, since mere part-of-speech tag sequences or chunk phrase sequences does not work. Rules for relations should be at the word level. We have five relationship types between treatment and problem, the semantics of the relations are very close that it becomes difficult in framing rules. The sentences with TrIP (improves) relationship type have verbs like improvement, relieved, resolved, controlled, etc., which are directly attributed to TrIP. Where as the relationship types TrCP (causes) and TrWP (worsens) are bit similar as both signifies the negative impact of Treatment on the Problem. For example, the pattern \\``$<problem> resistant~to <treatment>$'' was treated as TrCP, where as,\\ ``$<problem> intermittently~resistant~to <treatment>$'' was tagged as TrWP. 
Few sentences does not even have direct verbs to know which relationship type they belong to, for example, ''$<problem> of~the <treatment>$'', ``$<problem> as~he~was~on <treatment>$'', ``$<problem> since~the <treatment>$'', etc. Also we have found few ambiguous relationship annotations in the original dataset. In few sentences like ``Known allergies to Drugs'' are sometimes tagged as TrAP and TrCP in our dataset.
Few of the ambiguous patterns in gold standard dataset are listed below:
\begin{itemize}
\item ``$<treatment> concern~for <problem>$'' has given any of TrAP, TrIP or TrCP relationship types.
\item ``$<treatment> treated~for <problem>$'' can take both TrAP and TrIP
\item ``$<problem> after <treatment>$'' has both TrAP and TrIP relationship types.
\item Similarly, for other patterns like ``$<problem>due~to<treatment>$'', etc.
\end{itemize} 

\subsubsection{Performance of Hybrid Bi-LSTM and Rule-based system}
We merge the rule-based approach relationships with high precision to Bi-LSTM outcomes. That is, from Table~\ref{tab:results_rel_additionalfeatures}, except ``'TrAP', all other relationship samples of rule-based approach are combined with Bi-LSTM samples.  Table \ref{tab:sota-re} shows that this combination gave better performance for TrIP and TrNAP relationships. The numbers of TrCP and TrWP are unchanged since the samples of those relationship types were already identified by Bi-LSTM.
The source code of our system is available at:\\ https://github.com/RaghavendraCh/RelationExtraction\_keras. 

\pagebreak



\section{Conclusion}

In this paper, we have shown the important problem of extracting relationships between treatment and medical problem. We have described Bi-directional LSTM approach to automatically extract the relationships. The neural network is trained by considering both word-level and sentence-level representations. We have experimented Bi-LSTM by varying the number of negative samples, embedding vector sizes and features, by keeping the parameters ``lstm\_output\_size'' and ``number\_of\_epochs'' as constant. We have used support vector machines as the baseline system for relation extraction. Bi-LSTM has shown relatively better performance on identifying TrAP relation, implying the importance of a large number of data samples required to train a neural network model.  Also, the rule-based approach has been used for relationships with less number of samples available in our dataset. 
Finally, we have shown that rule-based system with high-precision leverages the performance of the neural network. As a future work, we plan to extend our work to different kinds of relationships, such as test-problem, drug-problem, and drug-drug relations, that exists in the biomedical literature.

\bibliographystyle{abbrv}
\bibliography{relations}

\end{document}